\documentclass{article}

\PassOptionsToPackage{numbers,sort}{natbib}

\usepackage{subcaption}
\usepackage{graphicx}
\usepackage[export]{adjustbox}

\usepackage[preprint]{neurips_2023}

\usepackage[utf8]{inputenc} 
\usepackage[T1]{fontenc}    
\usepackage{hyperref}       
\usepackage{url}            
\usepackage{booktabs}       
\usepackage{amsfonts}       
\usepackage{nicefrac}       
\usepackage{microtype}      
\usepackage{xcolor}         

\usepackage{mathtools}
\usepackage{algorithm2e}
\RestyleAlgo{ruled}

\usepackage{subcaption}

\DeclarePairedDelimiterX{\infdivx}[2]{(}{)}{%
  #1\;\delimsize\|\;#2%
}

\DeclarePairedDelimiter{\norm}{\lVert}{\rVert}

\title{Contrastive Language-Image Pretrained Models are Zero-Shot Human Scanpath Predictors}

\author{%
  Dario~Zanca\\
  FAU Erlangen-N\"urnberg\\
  91052 Erlangen, Germany \\
  \texttt{dario.zanca@fau.de} \\
  \And
  Andrea~Zugarini\\
  Expert.AI\\
  53100 Siena, Italy \\
  \texttt{azugarini@expert.ai} \\
  \And
  Simon~Dietz\\
  FAU Erlangen-N\"urnberg\\
  91052 Erlangen, Germany \\
  \texttt{simon.j.dietz@fau.de} \\
  \And
  Thomas R.~Altstidl\\
  FAU Erlangen-N\"urnberg\\
  91052 Erlangen, Germany \\
  \texttt{thomas.r.altstidl@fau.de} \\
  \And
  Mark A.~Turban Ndjeuha\\
  FAU Erlangen-N\"urnberg\\
  91052 Erlangen, Germany \\
  \texttt{mark.turban@fau.de} \\
  \And
  Leo~Schwinn\\
  FAU Erlangen-N\"urnberg\\
  91052 Erlangen, Germany \\
  \texttt{leo.schwinn@fau.de} \\
  \And
  Bjoern M.~Eskofier\\
  FAU Erlangen-N\"urnberg\\
  91052 Erlangen, Germany \\
  \texttt{bjoern.eskofier@fau.de} \\
  }

\begin{document}

\maketitle

\begin{abstract}
Understanding the mechanisms underlying human attention is a fundamental challenge for both vision science and artificial intelligence. While numerous computational models of free-viewing have been proposed, less is known about the mechanisms underlying task-driven image exploration. 
To address this gap, we present CapMIT1003, a database of captions and click-contingent image explorations collected during captioning tasks. CapMIT1003 is based on the same stimuli from the well-known MIT1003 benchmark, for which eye-tracking data under free-viewing conditions is available, which offers a promising opportunity to concurrently study human attention under both tasks. We make this dataset publicly available to facilitate future research in this field. 
In addition, we introduce NevaClip, a novel zero-shot method for predicting visual scanpaths that combines contrastive language-image pretrained (CLIP) models with biologically-inspired neural visual attention (NeVA) algorithms. NevaClip simulates human scanpaths by aligning the representation of the foveated visual stimulus and the representation of the associated caption, employing gradient-driven visual exploration to generate scanpaths.
Our experimental results demonstrate that NevaClip outperforms existing unsupervised computational models of human visual attention in terms of scanpath plausibility, for both captioning and free-viewing tasks. Furthermore, we show that conditioning NevaClip with incorrect or misleading captions leads to random behavior, highlighting the significant impact of caption guidance in the decision-making process. These findings contribute to a better understanding of mechanisms that guide human attention and pave the way for more sophisticated computational approaches to scanpath prediction that can integrate direct top-down guidance of downstream tasks. 
\end{abstract}

\section{Introduction}
Visual attention is an essential cognitive process in human beings, enabling selective processing of relevant portions of visual input while disregarding irrelevant information. This function not only facilitates efficient  information processing but also plays a crucial role in shaping perception and awareness \citep{kietzmann2011overt, macknik2009role}. By directing attention toward specific content within a scene, individuals are able to construct a coherent and meaningful representation of their visual environment. Seminal works by \citet{buswell1935people} and \citet{tatler2010yarbus} investigated the relationship between eye-movement patterns and high-level cognitive factors to demonstrate the influential role of task demands on eye movements and visual attention. According to \citet{tatler2005visual}, human eye movements follow a distinctive temporal pattern. Individuals' initial fixations are primarily influenced by bottom-up visual features, displaying a notable consistency across different subjects. However, these fixations rapidly diverge as individuals' unique goals and motivations, i.e. their tasks, come into play. This relationship between attention and task becomes particularly relevant when attention interplays with language: \citet{altmann2007real} suggested a link between language processing and visual attention based on object affordances, while \citet{mishra2009interaction} demonstrated the influence of visual attention on language behavior.

\begin{figure}[t]
    \centering
    \includegraphics{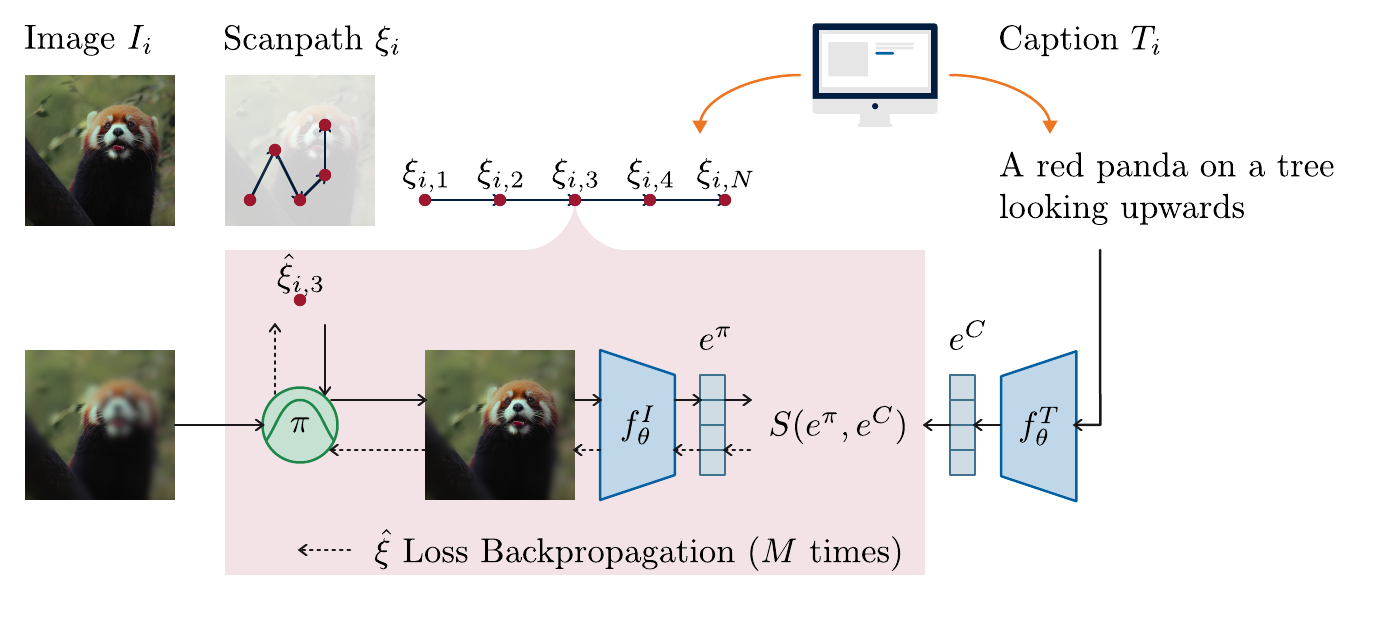}
    \caption{\textbf{Method Overview (NevaClip).} For each image $I_i$ a scanpath $\xi_i$ consisting of user clicks and a caption $T_i$ are collected in a web interface. To predict the next fixation, here $\xi_{i,3}$, NevaClip's attention mechanism combines past fixations $\{\hat{\xi}_{i,j} \mid j < 3\}$ with an initial guess for $\hat{\xi}_{i,3}$ to create a partially blurred image. This image is passed to CLIP's image embedding network $f_\theta^I$ to obtain an image embedding $e^\pi$. The cosine similarity loss $S(e^\pi,e^C)$ between $e^\pi$ and the text embedding $e^C$ obtained by passing the caption to CLIP's text embedding network $f_\theta^T$ is then computed as a measure of how well the current scanpath aligns with the given caption. By backpropagating this loss $M$ times, NevaClip finds the fixation $\hat{\xi}_{i,3}$ that minimizes this alignment loss.}
    \label{fig:overview}
\end{figure}

Despite great interest in understanding mechanisms underlying task-driven visual exploration, most current computational models of visual attention focus on free viewing \citep{borji2012state}, which primarily involves bottom-up saliency, and overlooks the influence of different tasks. In \citet{itti1998model}, conspicuity maps are combined to create a saliency map, which expresses for each location in the visual field the probability of being attended by humans. Recently, deep learning approaches demonstrated state-of-the-art performance in saliency prediction tasks \citep{pan2017salgan,cornia2018sam,wang2019revisiting}. Saliency methods can be combined with the Winner-Take-All algorithm~\citep{koch1987shifts}, to progressively generate fixation sequences. Gaze shifts are described in~\citep{boccignone2004modelling} as a stochastic process with non-local transition probabilities in a saliency field. A constrained random walk is performed in this field to simulate attention as a search mechanism. 
\citet{zanca2017variational} describe attention as a dynamic process, where eye movements are governed by three fundamental principles: boundedness of the retina, attractiveness of gradients, and brightness invariance. \citet{zanca2019gravitational} propose a mechanism of attention that emerges from the movement of a unitary mass in a gravitational field. In this approach, specific visual features, such as intensity, optical flow, and a face detector, are extracted for each location and compete to attract the focus of attention. Recently, \citet{schwinn2022behind} proposed the Neural Visual Attention (NeVA) algorithm to generate visual scanpaths in a top-down manner. This method imposes a biologically-inspired foveated vision constraint to generate human-like scanpaths without directly training for this objective but exploiting the top-down guidance of the loss of a pretrained vision model in an iterative optimization scheme.

To investigate task-driven human attention and  its interplays with language, in this paper we contribute computational models that simulate human attention during the captioning process, while additionally providing a dataset specifically designed for this task. Generating accurate and relevant descriptions of images \citep{li2019visual,sharma2020image} is an essential task that requires a deep understanding of both the human visual attention mechanism and its connection to language processing. Numerous studies have explored the characteristics of soft attention mechanisms in captioning models, focusing on their correlation with objects present in the scene and their consistency with human annotations \citep{liu2017attention, mun2017text}. Other works demonstrated the effectiveness of incorporating computational models of visual attention into captioning models \citep{chen2018boosted}. However, to date, there is a lack of computational models that accurately capture the complexities of human attention during captioning tasks and flexibly adapt to multimodal language-image sources. 

First, we collect behavioral data from human subjects using a web application. With this procedure, we create CapMIT1003, a database of captions and click-contingent image explorations collected under captioning tasks. CapMIT1003 is based on the well-known MIT1003 dataset \citep{judd2009learning}, which comprises 1003 images along with corresponding eye-tracking data of subjects engaged in free-viewing. We developed a click-contingent paradigm, drawing inspiration from prior studies on free viewing \citep{jiang2015salicon}, which is designed to emulate the natural viewing behavior of humans. By exploiting a computer mouse as a substitute for an eye tracker, this approach enabled a large-scale data collection, without the need for specialized equipment. This integrated dataset holds significant importance as it enables the study of both tasks simultaneously. We make this dataset publicly available as a valuable resource for investigating the relationship between human attention, language, and visual perception, fostering a deeper understanding of the cognitive processes involved in captioning.

To study the relationships between the collected visual explorations and  captions, we combine the previously illustrated Neural Visual Attention \citep{schwinn2022behind} algorithms with Contrastive Language-Image Pretrained (CLIP) \citep{radford2021learning} models, that effectively map images and associated captions onto the same feature representation. The basic idea is that, by selectively attending to and revealing portions of the image that are semantically related to the concepts expressed in the caption, we increase the cosine similarity between the visual embedding of the foveated image and its corresponding caption. An overview of the approach is illustrated in Figure \ref{fig:overview}. By setting this as a target for the NeVA algorithm, subsequent fixations are chosen to maximize the cosine similarity between the visual and text embeddings through an iterative optimization process. It is worth noticing that this method is entirely unsupervised, as no human data is utilized for training. 

Our findings demonstrate that generating scanpaths conditioned on the correctly associated captions results in highly plausible scanpath trajectories, as evidenced by the similarity to corresponding human attention patterns. When conditioning the scanpath generation on captions provided by other users for the same image, scanpath plausibility slightly decreases. However, generating scanpaths conditioned on random captions, given by other subjects on different images, leads to systematically unplausible visual behaviors, with performance equivalent to that of a random baseline. Furthermore, we observe that, by employing visual embedding of the clean image as a target caption representation, we achieve the best results in terms of scanpath plausibility, setting a state-of-the-art result for the newly collected dataset. These findings together highlight the effectiveness of our approach in leveraging captions to guide visual attention during scanpath generation and demonstrate the utility of contrastive language-image pretrained models as zero-shot scanpath predictors. All data and code will be released on GitHub after publication.

\section{Data collection}
In this section, we introduce CapMIT1003, an integrated dataset combining captions, and click-contingent image explorations. This dataset is crucial as it  offers a valuable resource to explore the interplay between human attention, language, and visual perception, contributing to a deeper understanding of cognitive processes in captioning. It builds over the well-known benchmark MIT1003 \citep{judd2009learning}, which contains eye-tracking data collected under free-viewing conditions.

\subsection{Web-based crowdsourcing of visual explorations and captions}
We developed a web application that presents images to participants and prompts them to provide textual descriptions while performing click-contingent image explorations. Users were asked to visually explore and provide a caption for up to fifty distinct images from the MIT1003 dataset \citep{judd2009learning}. Images were presented sequentially, one after another. We used the protocol defined in \citet{jiang2015salicon} for which crowdsourcers are presented with a completely blurred version of the image, and can unveil details on specific locations by clicking. A click will reveal information in an area corresponding to two degrees of visual angle, calculated with respect to the original dataset experimental setup described in \citep{judd2009learning}. This process is aimed at simulating a human's foveated vision. Subjects can click up to ten times, before providing a caption. We instructed users to describe the content of the image with "one or two sentences", in English. All clicks and captions are stored in a database, while no personal information has been collected.

\subsection{CapMIT1003 statistics}

\begin{figure}[!tbp]
  \centering
  \begin{minipage}[b]{0.43\textwidth}
    \includegraphics[width=\textwidth]{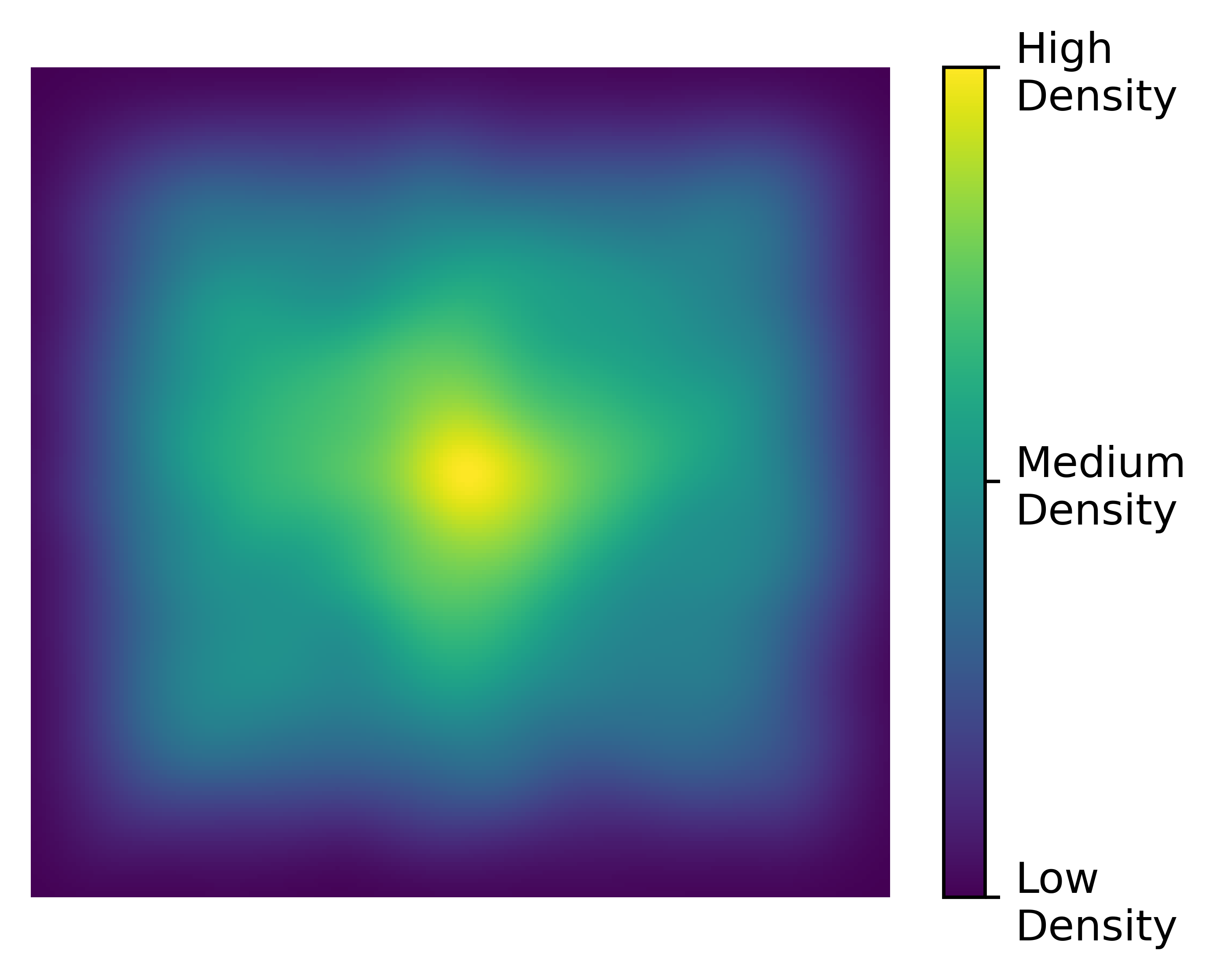}
    \captionof{figure}{Estimated clicks density.\newline}
    \label{fig:density}
  \end{minipage}
  \hfill
  \begin{minipage}[b]{0.49\textwidth}
    \includegraphics[width=\textwidth]{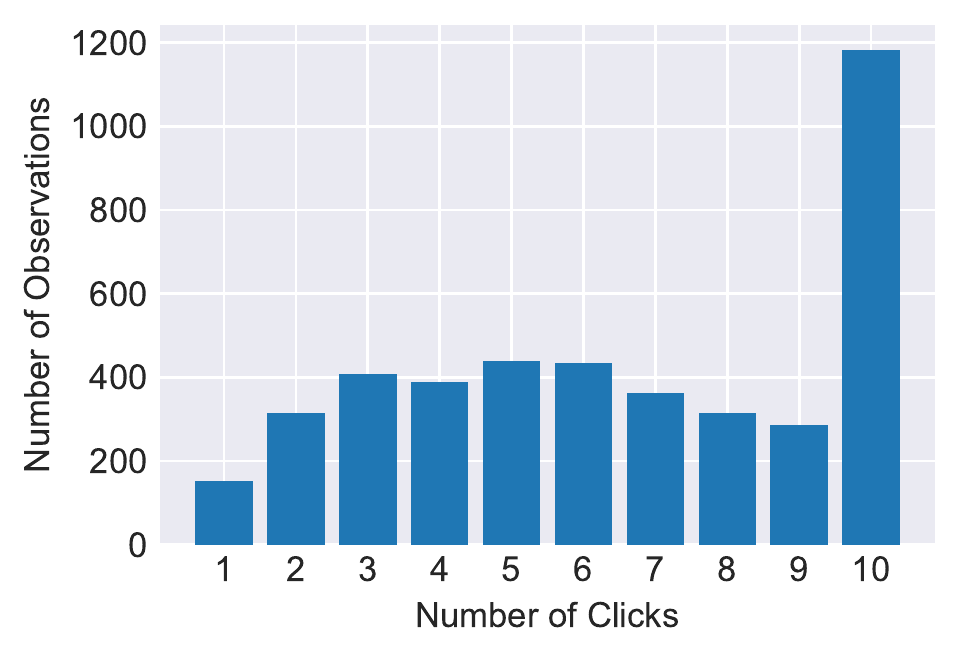}
    \captionof{figure}{Number of observations with different numbers of clicks.}
    \label{fig:click_count}
  \end{minipage}
\end{figure}

In total, we collected 27865 clicks on 4573 observations over 153 sessions.
We excluded 654 observations that were skipped, 33 with no recorded clicks, and 38 captions with less than 3 characters. After these exclusions, we analyzed the data for 1002 of the 1003 images in the pool.
The average number of characters per caption was 50.89, with a standard deviation of 33.88. The average number of words per caption was 9.78, with a standard deviation of 6.47. The maximum caption lengths were 259 characters and 51 words, respectively. Figure \ref{fig:density} shows the click density based on all collected clicks. The figure was generated by scaling the position of each click with the size of the corresponding image. Afterward, we used a Gaussian kernel density estimation to calculate the click density. Figure \ref{fig:click_count} shows the number of observations with each click count from 1 to 10. Using all 10 available clicks was the most favored option. 

\section{Method}

\subsection{CLIP}
Contrastive Language-Image Pretraining (CLIP) \cite{radford2021learning}, is a method of learning semantic concepts efficiently from natural language supervision, which is trained on a large dataset of $N$ (image, text) pairs. CLIP uses a \textit{shared} embedding space for both modalities. Let $f_{\theta}^{I}$ be the image encoder and $f_{\theta}^{T}$ be the text encoder, where $\theta$ represents the trainable parameters of the models. Given an image $I_i$ and a corresponding text $T_i$, the image embedding is defined as $e_i^I=f_{\theta}^{I}(I_i)$, and the text embedding is defined as $e_i^T=f_{\theta}^{T}(T_i)$. 

CLIP models are trained by maximizing the cosine similarity of the embeddings of the $N$ correct pairs in the batch while minimizing the cosine similarity of the embeddings of the $N^2-N$ incorrect pairings. The cosine similarity between two embeddings $e_i$ and $e_j$ is computed as $S_c(A, B) = (A \cdot B) / (\norm{A} \norm{B})$.

\subsection{Neural Visual Attention (NeVA)}
Let $\xi = \{\xi_t\}_{t \in \{1, ..., N\}}$ be a sequence of fixation points, also called \textit{scanpath} in what follows. Given a visual stimulus $S$ and its coarse version $\Tilde{S}$, \citet{schwinn2022behind} define a foveated stimulus as $$\pi \left(S, \xi_t \right) = G_{\sigma_\xi}(\xi_t) \cdot S + (1-G_{\sigma_\xi}(\xi_t)) \cdot \Tilde{S},$$ 
where $\xi_t$ is the current fixation point at time $t$, and $G_{\sigma_\xi}$ is a Gaussian blob centered on $\xi_t$ with standard deviation $\sigma_\xi$. A scanpath is generated iteratively, by an optimization procedure. Let $\tau$ be any task model, and $\mathcal{L}_{\tau}$ its associated loss. The next fixation location $\xi_{t+1}$ is determined by a local search in order to minimize the loss 
$$\mathcal{L}_{\tau} (\tau(\pi \left(S, \xi_{t+1} \right), y)), $$
where $y$ is a target output for the specific downstream task. This approach can be applied to any pretrained vision model.

\SetKwComment{Comment}{/* }{ */}
\SetKwInOut{Parameter}{Parameters}
\SetKwInOut{Variables}{Variables}
\SetKwInOut{Output}{Output}
\newcommand\mycommfont[1]{\footnotesize\ttfamily\textcolor{blue}{#1}}
\SetCommentSty{mycommfont}
\begin{algorithm}
\caption{Pseudocode illustrating the NeVAClip optimization algorithm}\label{alg:nevaclip}
\Parameter{Number of fixations: $N$, Visual stimulus: $S$, Blurred visual Stimulus: $\Tilde{S}$\\Foveation size: $\sigma_{\xi}$, Optimization steps $M$, Caption embedding $e^C$\\Learning rate: $\alpha$}
\BlankLine
\Variables{Clip visual encoder: $f_{\theta}^{I}$, Gaussian blop: $G_{\sigma_\xi}$, Gradient update: $\delta$}
\BlankLine

$\xi \gets \text{list}((N, 2))$  \tcp*{initialize a list of fixations centered in the image}

\For{$t\gets0$ \KwTo $N$}{
    
    $\mathcal{L}_{best} \gets \infty$ \tcp*{track lowest loss found}
    
    $\xi_{best} \gets (0, 0)$ \tcp*{track fixation position corresponding to lowest loss}
    
    \For{$m\gets0$ \KwTo $M$}{
        $G_{\sigma_\xi}(t) \gets \mathcal{N}\left(\xi_t^0, \sigma_{\xi}\right)$ \tcp*{calculate gaussian blob}
    
        $\pi \left(S, \xi_t^m \right) \gets G_{\sigma_\xi}(t) \cdot S + (1-G_{\sigma_\xi}(t)) \cdot \Tilde{S}$ \tcp*{calculate foveated stimulus}

        $e^{\pi} \gets f_{\theta}^{I}\left(\pi\left(S,\xi_t^m\right)\right)$ \tcp*{calculate embedding for given stimulus}
        
        $\mathcal{L}_t^m \gets S_c\left(e^\pi, e^C\right)$ \tcp*{calculate cosine similarity between embeddings}

        $\delta \gets \nabla \mathcal{L}\left(\xi_t^m; \pi\right)$ \tcp*{calculate loss gradient}
        
        $\xi_t^{m + 1} = \xi_t^m + \alpha \cdot \nabla \mathcal{L}\left(\xi_t^m; \pi\right)$ \tcp*{update foveation position}

        \uIf{$\mathcal{L}_t^m < \mathcal{L}_{best}$}
        {
            $\mathcal{L}_{best} \gets \mathcal{L}_t^m$ 

            $\xi_{best} \gets \xi_t^m$
        }
    }
    $\xi_t \gets \xi_{best}$ \tcp*{save fixation position corresponding to lowest loss}
}
\Output{Scanpath: $\xi$}
\end{algorithm}

\subsection{Generating scanpaths with CLIP embeddings (NevaClip)}
To use NeVA in combination with CLIP, we need to modify the optimization procedure to incorporate the cosine similarity between (foveated) stimuli and caption embeddings in the loss function. 

Let $S$ be a visual stimulus and $C$ be its associated caption. We use CLIP's visual encoder $ f_{\theta}^{I}$ as the task model, where the downstream task is, at each step, to maximize the cosine similarity between the foveated image embedding and caption embeddings, i.e., 
\begin{equation}\label{eq:cosinesim}
    S_c\left(e^{\pi}, e^C\right),
\end{equation}
where $e^{\pi} = f_{\theta}^{I}\left(\pi \left(S, \xi_t \right)\right)$, and $e^C = f_{\theta}^{T}\left(C\right)$.
Then, the NeVA optimization procedure described in the previous subsection can be extended to generate a complete visual scanpath for a given stimulus and its associated caption embedding $e^C$. To simplify the notation, we denote this procedure as 
\begin{equation}
\label{eq:standard}
NevaClip(S, e^C) = \xi^{S,C},
\end{equation}
where $\xi^{S,C}$ is the sequence of fixation points that constitute the scanpath. This approach provides an efficient method for generating scanpaths that are optimized to minimize the distance between the foveated versions of the stimulus, highlighting specific portions of the image, and the semantic concepts included in the captions. The pseudo-code in algorithm \ref{alg:nevaclip} illustrates the overall procedure.

We observe that, as a consequence of its training, the CLIP visual embedding of the clean image $e^S = f_{\theta}^{I}\left( S \right)$ (i.e., not foveated), can be considered as the \textit{ideal} caption's embedding. By employing the CLIP visual embedding of the clean image as a target in the NeVA optimization procedure, we can generate a scanpath without an explicit caption, i.e., 
\begin{equation}
\label{eq:fullyvisual}
    NevaClip(S, e^S) =  \xi^S.
\end{equation}
We refer to this approach as the  \textit{NevaClip (\textit{visually-guided})}. Although it violates the biological constraint of partial information by having access to the whole stimulus (we in fact need the entire clean image in advance to compute $e^S$), this serves us as a gold standard as it potentially minimizes the distribution shift between captions provided by our users in our study and captions that CLIP would assign the same image. Furthermore, this configuration of the NevaClip model is relevant for practical applications in which processing the entire stimulus is feasible, while we do not have real-time access to the subject's caption.

\begin{figure}[t]

  \centering
  
  \begin{subfigure}[t]{0.33\textwidth}
  \includegraphics[width=\linewidth,valign=t]{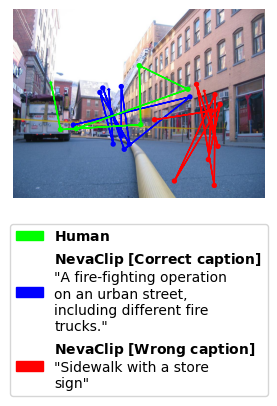}
  \end{subfigure}\hfill
  \begin{subfigure}[t]{0.33\textwidth}
  \includegraphics[width=\linewidth,valign=t]{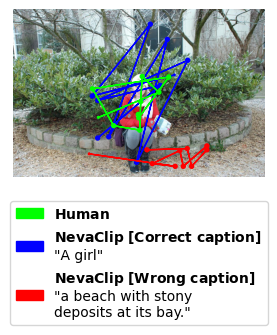}
  \end{subfigure}\hfill
  \begin{subfigure}[t]{0.33\textwidth}
  \includegraphics[width=\linewidth,valign=t]{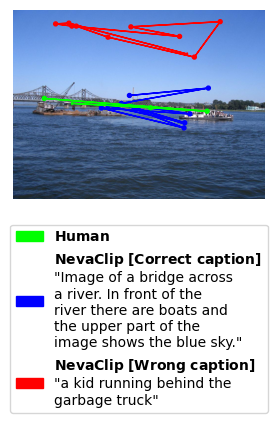}
  \end{subfigure}
  \caption{\textbf{Scanpaths comparison. }The figure shows three images along with the scanpaths. The Human scanpath represents the actual sequence of clicks performed by a human observer (green). The other two scanpaths are generated with NevaClip by using the correct caption provided by the same subject (blue), or a wrong caption (red) given by a different subject on a different image.
}
  \label{fig:scanpaths_examples}
\end{figure}

\section{Experiments}

\subsection{Experimental setup}
We adhere to the original implementation of the Neva algorithm with 20 optimization steps\footnote{\url{https://github.com/SchwinnL/NeVA}}. Random restarts during the local search were not allowed, as we found out this induces a favorable proximity preference \citep{koch1987shifts} to the next fixation.  In \citep{schwinn2022behind} past explorations are incorporated into the agent's state, modulated by  a forgetting factor. In our experiments, we found that optimal results were achieved for NevaClip with a forgetting factor of $0$. This indicates that the system operates under a Markovian assumption, where only information from the last fixation is used in order to determine the next location to be attended. We use the implementation of the CLIP model provided by the original authors\footnote{\url{https://github.com/openai/CLIP}}. When not stated otherwise, all NevaClip results are presented for the Resnet50 backbone.  For all  competitor models, the original code provided by the authors is used. 

\subsection{Baselines and models}
In order to investigate the relationship between captions and their corresponding visual explorations, we evaluate four distinct variations of our NevaClip algorithm. These versions differ only on the target embedding used as a guide for visual exploration, as described in equations~\ref{eq:cosinesim}, \ref{eq:standard}, and \ref{eq:fullyvisual}.

\textbf{NevaClip (correct caption).} Scanpaths are generated maximizing the alignment between visual embeddings of the foveated stimuli, and text embeddings of the corresponding captions, as described in equation~\ref{eq:standard}.\\
\textbf{NevaClip (different caption, same image). }Scanpath are generated to maximize the alignment between visual embeddings of the foveated stimuli, and text embeddings of the captions provided by a different subject on the same image.\\
\textbf{NevaClip (different caption, different image).} Scanpath are generated to maximize the alignment between visual embeddings of the foveated stimuli, and text embeddings of the random captions provided by a different subject on a different image.\\
\textbf{NevaClip (visually-guided).} Scanpath are generated to maximize the alignment between visual embeddings of the foveated stimuli, and the visual embedding of the clean version of the same image, as described in equation~\ref{eq:fullyvisual}.

We define three baselines, to better position the quality of the results.

\textbf{Random.} Scanpaths are generated by subsequently sampling fixation points from a uniform probability distribution defined over the entire stimulus area.\\
\textbf{Center.} Scanpaths are randomly sampled from a $2$-dimensional Gaussian distribution centered in the image. We used the center matrix provided in \citet{judd2009learning}, associated with the MIT1003 dataset.\\
\textbf{Clicks Density.} Scanpaths are randomly sampled from a $2$-dimensional density distribution obtained by cumulating all clicks in the dataset.

We compare our approach to four state-of-the-art unsupervised models of human visual attention. It is worth mentioning that none of these models is designed to exploit information from the caption.

\textbf{NeVA (original)}~\citep{schwinn2022behind}. We employ the guidance of a classification downstream task, which was demonstrated to produce the highest similarity to humans in the original paper. We use the optimized version of the algorithm, provided by the authors.\\
\textbf{Constrained Levy Exploration (CLE)}~\citep{boccignone2004modelling}. Scanpath is described as a realization of a stochastic process. We used saliency maps by~\citep{itti1998model} to define non-local transition probabilities. Python implementation provided by the authors in~\citep{boccignone2019look}.\\
\textbf{Gravitational Eye Movements Laws (G-EYMOL)}~\citep{zanca2019gravitational}. Scanpath is obtained as the motion of a unitary mass within gravitational fields generated by salient visual features.\\
\textbf{Winner-take-all (WTA)}~\citep{koch1987shifts}. Based on the saliency maps by~\citet{itti1998model}, the next fixation is selected at the location of maximum saliency. Inhibition-of-return allows switching to new locations.

\begin{table}[t]
\caption{\textbf{Average ScanPath Plausibility ($SPP$).} We summarise the SPP SBTDE scores for all baselines, competitors, and NevaCLIP versions. The scores are computed for each sublength $s \in \{1, ..., 10\}$, and the average of mean and standard deviation for all sublength is presented. Best in \textbf{bold}, second best \underline{underlined}.}
\centering
\begin{tabular}{lcc}
\hline
\textbf{} & \textbf{CapMIT1003} & \textbf{MIT1003} \\
\textbf{Model} & Captioning & Free-viewing \\
\hline
\hline
Random baseline & 0.26 (0.13) & 0.30 (0.11)  \\
Center baseline & 0.27 (0.13) & 0.33 (0.12) \\
ClickDensity baseline & 0.27 (0.13) & 0.32 (0.11) \\
\hline
G-Eymol & 0.34 (0.17) & 0.46 (0.16) \\
Neva (original) & 0.34 (0.17)
 & 0.47 (0.16) \\
CLE & 0.30 (0.19)
 & 0.47 (0.18) \\
WTA+Itti & 0.31 (0.16) & 0.37 (0.14) \\
\hline
NevaClip (correct caption) & \underline{0.35} (\underline{0.18}) & \underline{0.50} (\underline{0.17}) \\
NevaClip (different caption, same image) & 0.34 (0.17) & \underline{0.50} (\underline{0.17}) \\
NevaClip (different caption, different image) & 0.26 (0.14) & 0.37 (0.15) \\
NevaClip (visually-guided) & \textbf{0.38 (0.19)} & \textbf{0.57 (0.18)} \\
\hline
\end{tabular}
\label{tab:all_results}
\end{table}

\subsection{Metrics and evaluation}
To measure the similarity between simulated and human scanpaths, we compute ScanPath Plausibility (SPP)~\cite{fahimi2021metrics}, using the string-based time-delay embeddings (SBTDE)~\cite{schwinn2022behind} as a basic metric. The SBTDE accounts for stochastic components which are typical of the process of human visual attention~\cite{pang2010stochastic}, by searching for similar subsequences at different times. This metric can be computed for each subsequence length, expressively indicating the predictive power of models at each stage of the exploration. The SBTDE at subsequence length $k$ is computed as 

$$\textit{SBTDE}_k(\xi^{A}, \xi^{B}) = \frac{1}{N} \sum_{x \in \Xi^{A}_k} d(x, \Xi^{B}_k),$$

where $\xi^{A}, \xi^{B}$ are the two scanpaths to be compared, and $\Xi^{A}_k, \Xi^{B}_k$ are the sets of all possible contiguous sublengths of size $k$ in $\xi^{A}$ and $ \xi^{B}$, respectively. We use the string-edit distance to compute the distance $d(x, \Xi^{B}_k)$ between subsequences.
The SPP accounts for large variability between human subjects, by considering only the human scanpath with the minimum distance for each stimulus. To obtain a similarity measure, ranging from $0$ (minimum similarity) and $1$ (maximum similarity), we consider the SBTDE complement to one. Let $\xi^S$ be a simulated scanpath, and $\Xi^H$ a set of human reference scanpaths, for each sublength $k$, scanpath plausibility is defined as 
$$SPP_k(\xi^S, \Xi^H) = max_{\xi^H \in \Xi^H} \left( 1 - SBTDE_k(\xi^S, \xi^H)\right).$$
Since human explorations, both for clicks and eye-tracking, can be of different lengths, for each subject we compute the metric up to $min(L, 10)$, where $L$ is the length of the human scanpath.

\begin{figure}[t]
    \centering
    \begin{subfigure}{0.49\textwidth}
        \includegraphics[width=\linewidth]{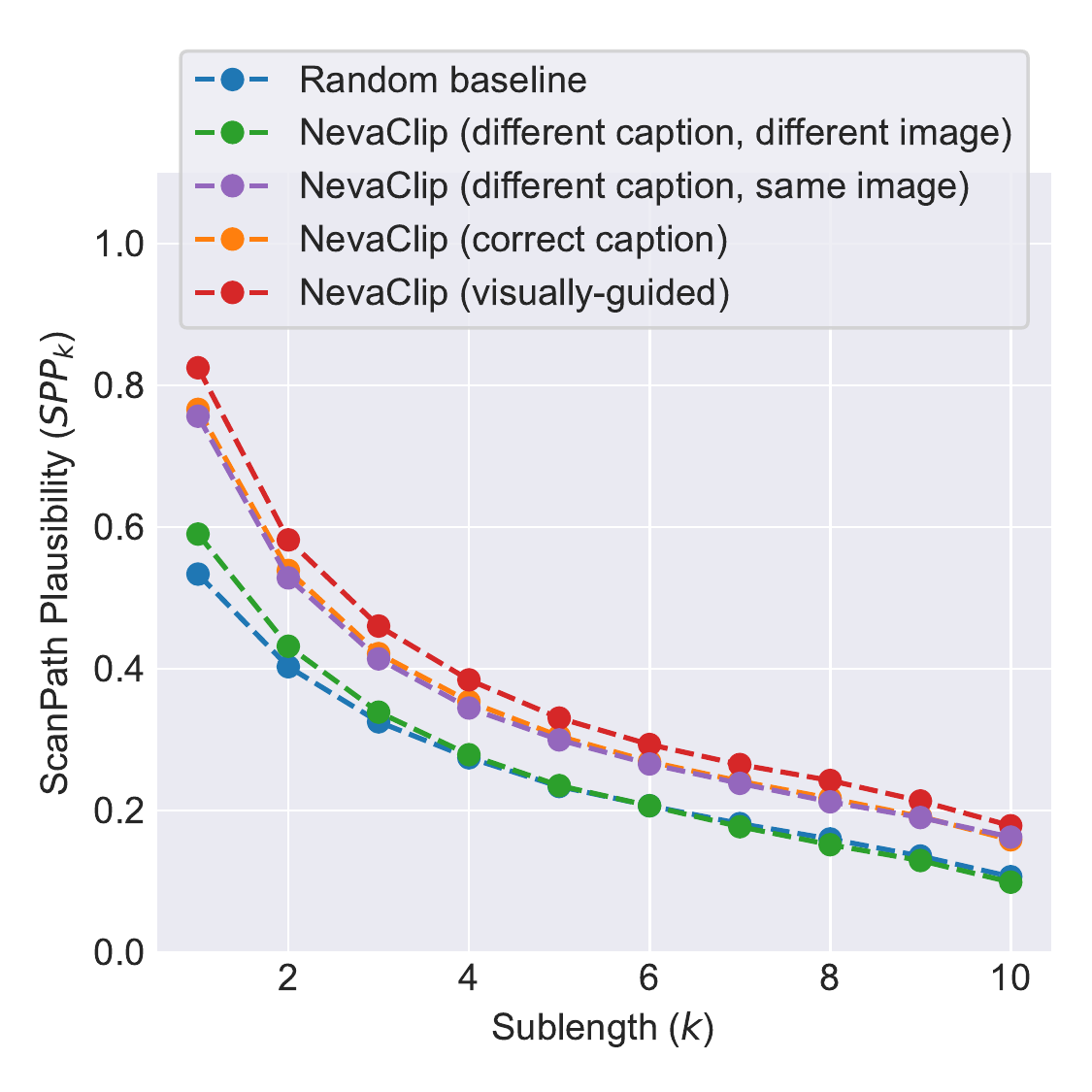}
        \caption{CapMIT1003 (captioning)}
        \label{fig:clicks}
    \end{subfigure}
    \hfill
    \begin{subfigure}{0.49\textwidth}
        \includegraphics[width=\linewidth]{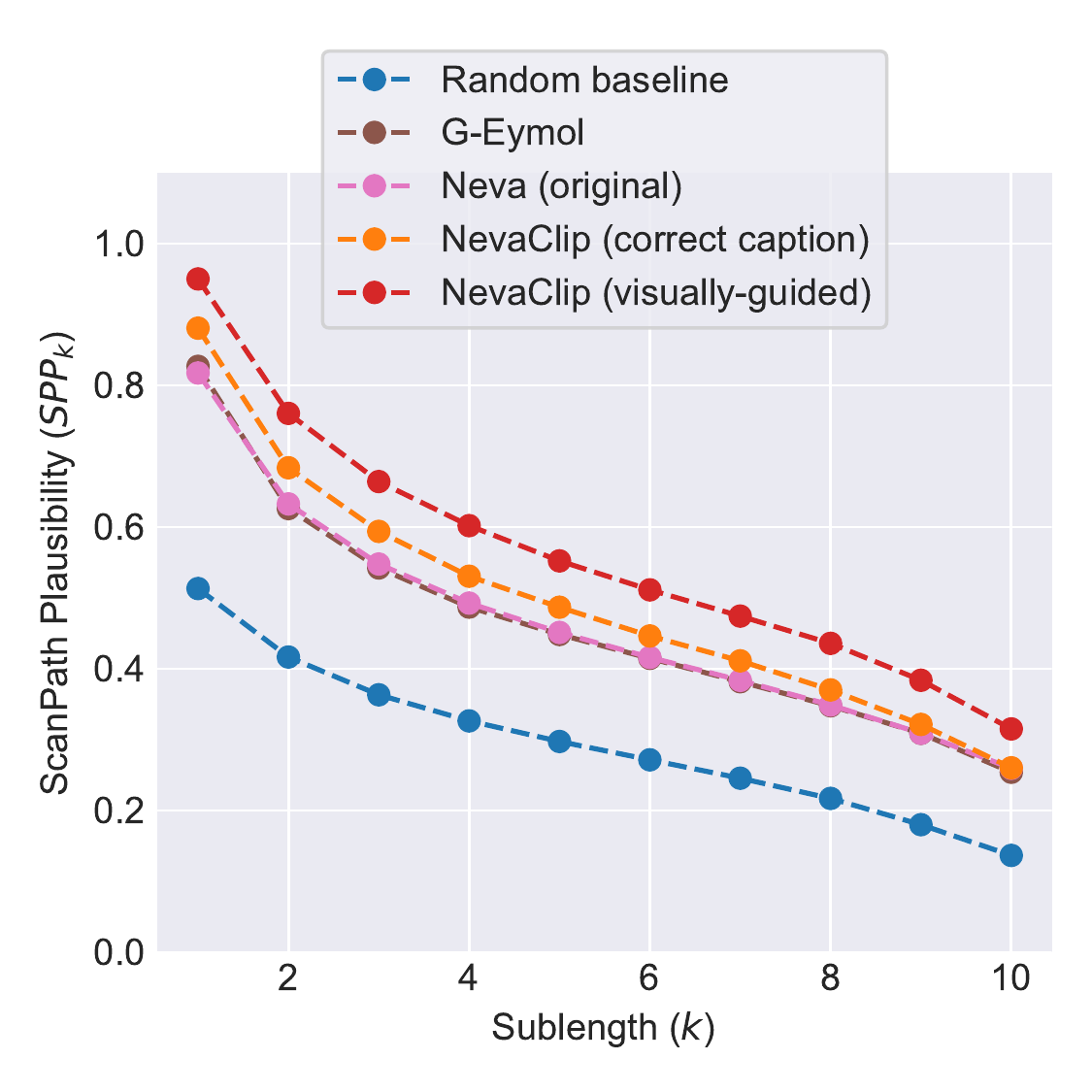}
        \caption{MIT1003 (free-viewing)}\label{fig:et}
    \end{subfigure}
    \caption{\textbf{ScanPath Plausibility ($SPP_K$) for each sublength. }Average SPP is plotted for each sublength $k \in \{1, ..., 10\}$. Fig. (a) shows a comparison of the different NevaClip configurations for the captioning task. Fig. (b) shows a comparison of NevaClip (correct caption) and NevaClip (visually-guided) with state-of-the-art unsupervised models G-Eymol and NeVA (original). }
    \label{fig:metrics}
\end{figure}

\subsection{Results on CapMIT1003 (captioning task)}

Table \ref{tab:all_results} summarises results on the CapMIT1003 dataset. The \textit{NevaClip (visually-guided)}, which exploits the guidance of the clean image embedding, significantly outperforms all other approaches by a substantial margin. As expected, \textit{NevaClip (different subject, different image)}, which generates a scanpath using a label from a different image and a different subject, performs similarly to the Random Baseline. We observed that this configuration systematically explores non-relevant objects that align with the misleading caption or, in their absence, emphasizes the background, which typically possesses an average feature representation (see Figure \ref{fig:scanpaths_examples}). The \textit{NevaClip (correct caption)}  performs slightly better than the \textit{NevaClip (different subject, same image)}, demonstrating that the caption provided by the subject brings useful information about their own visual exploration behavior. Among competitors, \textit{G-Eymol} and \textit{Neva (original)} compete well, despite not incorporating any caption information. It is worth noting that all approaches are unsupervised, and the scanpath prediction is performed  zero-shot. Supervised approaches could benefit from exploiting more information contained in the caption for personalized scanpath prediction.

\subsection{Results on MIT1003 (free-viewing task)}

The \textit{NevaClip (visually-guided)} approach again significantly outperforms all competitors by a substantial margin. \textit{NevaClip (correct caption)} and \textit{NevaClip (different subject, same image)} also performs better than state-of-the-art models \textit{NeVA (original)} and \textit{G-Eymol}. These results demonstrate a substantial overlap between the two tasks, as discussed later in section \ref{sec:captioningVSfreeviewing}. 

\subsection{The influence of CLIP's vision backbone.} We compare the performance of \textit{NevaClip (correct caption)} for different backbones. Results are presented in table \ref{tab:backbones}.  \textit{ResNet50} obtains a SPP SBTDE of 0.35 ($\pm$ 0.18), while \textit{ResNet101} and \textit{ViT-B/32} obtain 0.33 ($\pm$ 0.17) and 0.33 ($\pm$ 0.18), respectively. While the performance of different backbones is comparable, simpler models such as \textit{ResNet50} might benefit from simpler representations, which lead to less over-specific visual explorations. A similar observation was made by  \citet{schwinn2022behind}, where a combination of a simpler model and task exhibited the best performance in terms of similarity to humans.

\subsection{Critical comparison: free-viewing vs. captioning} \label{sec:captioningVSfreeviewing}

As expected, capturing attention in captioning tasks proves to be more challenging compared to free-viewing, as it involves the simultaneous consideration of multiple factors such as image content, language, and semantic coherence. This is reflected in the results, where all approaches achieve lower performance in terms of scanpath plausibility (see table \ref{tab:all_results}) for the captioning task. The ranking of models remains consistent across both tasks, with free-viewing models performing reasonably well on both datasets. This observation suggests a substantial overlap between free viewing and captioning, possibly due to the initial phases of exploration being predominantly driven by bottom-up factors rather than task-specific demands.

Simulating realistic scanpaths becomes increasingly challenging after just a few fixations, and this difficulty is particularly pronounced in the context of captioning, as illustrated in Figure \ref{fig:metrics}. This result is in line with existing literature \citep{tatler2005visual} demonstrating how human visual attention diverges quickly due to individual goals.

Upon analyzing the click density map depicted in Figure \ref{fig:density}, we observe that captioning elicits significantly more diverse exploratory patterns compared to free viewing, which exhibits a more pronounced center bias \citep{judd2009learning}. The same figure also reveals that a portion of the clicks is distributed in a grid-like fashion, indicating a tendency of users to perform a systematic  strategy for spread out visual exploration. This finding makes sense considering the nature of captioning, which assumes a holistic understanding of the scene, unlike free viewing which does not impose such requirements.

\begin{table}[t]
\setlength{\abovecaptionskip}{5pt}
    \caption{\textbf{Average ScanPath Plausibility ($SPP$) for different backbones.} We compare the performance in terms of scanpath plausibility of the \textit{NevaClip (correct caption)} approach, for different vision backbones. Best in \textbf{bold}.}\label{tab:backbones}
    \centering
    \begin{tabular}{lcc}
    \hline
    \textbf{} & \textbf{} &  \textbf{CapMIT1003}  \\
    \textbf{Backbone} & \textbf{Num. parameters} & Captioning  \\
    \hline
    \hline
    RN50 \citep{radford2021learning} & 26M & \textbf{0.35} (\textbf{0.18})  \\
    RN101 \citep{radford2021learning} & 44.6M & 0.33 (0.17) \\
    ViT-B/32 \citep{radford2021learning} & 86M & 0.33 (0.18) \\
    \hline
    \end{tabular}

\end{table}

\section{Conclusions}

In this paper, we first introduced the CapMIT1003 database, which consists of captions and click-contingent image explorations collected under captioning tasks. This dataset extends the well-known MIT1003 benchmark, collected under free-viewing conditions. We make this dataset publicly available to facilitate future research in this area. To model the relationship between scanpaths and captions, we introduced NevaClip, a novel zero-shot method for predicting visual scanpaths. This approach generates gradient-driven visual explorations under the biological constraint of foveated vision that demonstrate superior plausibility compared to existing unsupervised computational models of human visual attention, for both captioning and free-viewing tasks.

Our experimental results emphasized the significant impact of caption guidance in the decision-making process of NevaClip. Conditioning NevaClip with incorrect or misleading captions led to random behavior, highlighting the role of accurate caption information in guiding attention. This finding underlines the importance of future work in considering top-down guidance from downstream tasks in scanpath prediction and motivates the development of more sophisticated computational approaches.

\bibliographystyle{apalike} 
\bibliography{bibliography}

\newpage
\appendix
\section{Data Collection Instructions and Web Application Interface}

The following instructions were displayed to the subjects right before starting the experiment:

\begin{itemize}
    \item In this study you are asked to provide a caption for 50 images.
    \item Images are shown to you sequentially, one after another.
    \item Each image is initially completely blurry, and you can click on it up to ten times to reveal more details.
    \item After completing your exploration, describe the content of the image in 1-2 sentences in the text box on the right and click on submit. The caption must be in English.
    \item If you are very unsure of the content of the image, you can skip it by clicking the “Skip” button.
    \item Please take your time and try your best to reveal as much of the image as possible before providing a caption.
\end{itemize}

\begin{figure}[htbp]
  \begin{subfigure}[t]{0.45\textwidth}
    \centering
    \includegraphics[width=\linewidth]{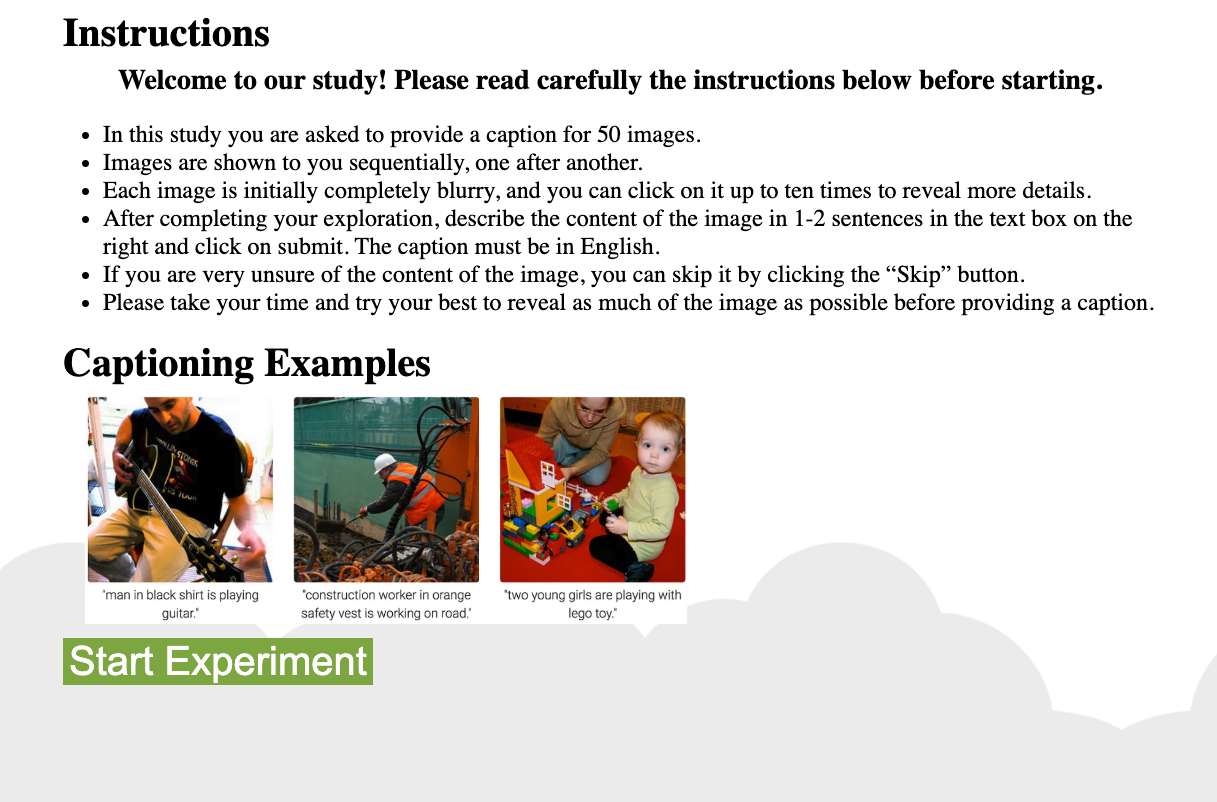}
    \caption{\textbf{Web application's instruction page.}}
    \label{fig:webapp_instructions}
  \end{subfigure}
  \hfill
  \begin{subfigure}[t]{0.45\textwidth}
    \centering
    \includegraphics[width=\linewidth]{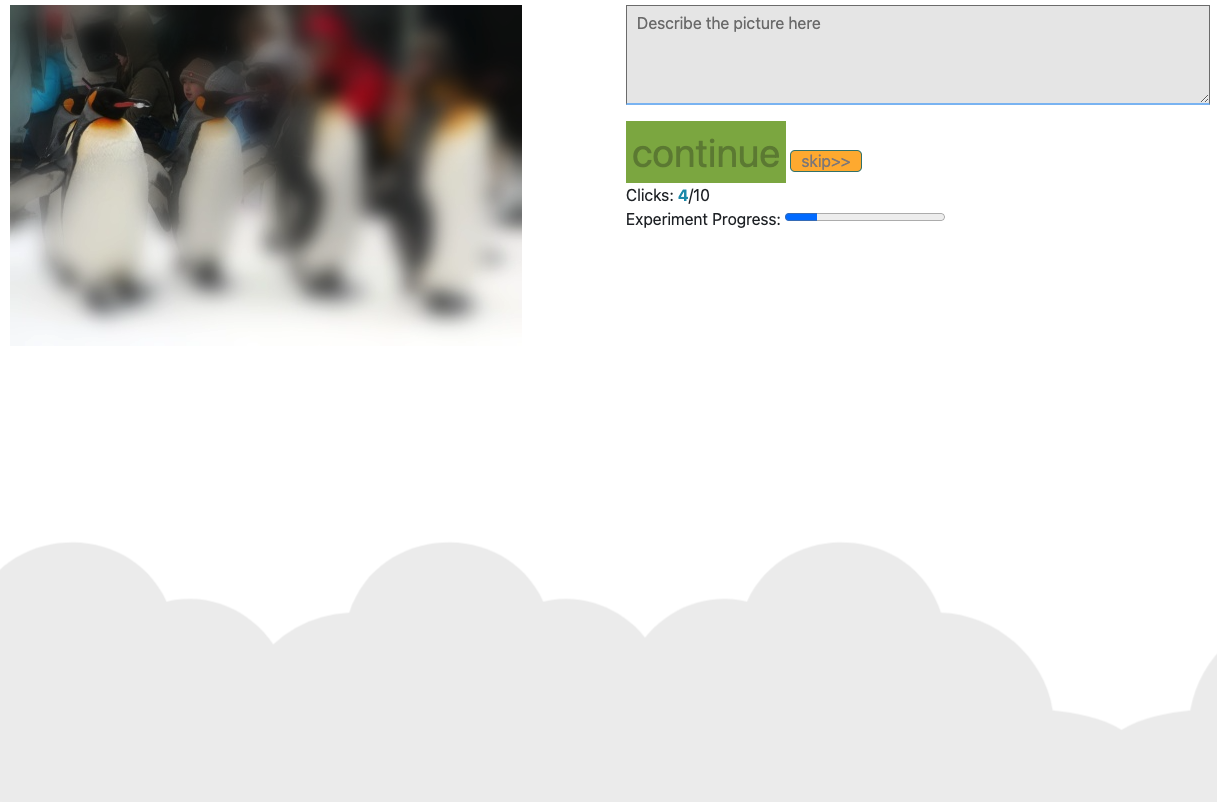}
    \caption{\textbf{Web application's captioning page.}}
    \label{fig:webapp_captioning}
  \end{subfigure}
  \caption{Web application screenshots.}
  \label{fig:webapp_screenshots}
\end{figure}

The web application's instruction page for data collection is depicted in Figure \ref{fig:webapp_instructions}. This page serves to acquaint participants with the task requirements by presenting them with examples of captions. Figure \ref{fig:webapp_captioning} showcases the appearance of the web application during the captioning task. The image is positioned on the left side, where clicking on it reveals details about the left-most penguin. On the right side, a text box is provided for participants to enter their captions after completing the exploration. Two buttons are available for participants: a "continue" button, which submits the provided caption, and a "skip" button, which can be used when participants are unable to provide a textual description.

\newpage
\section{Estimated Clicks Densities Split by their Order}

\begin{figure}[htbp]
\captionsetup[subfigure]{labelformat=empty}

\begin{subfigure}[t]{0.195\textwidth}
    \centering
        \caption{\textbf{1st}}
    \includegraphics[width=\linewidth]{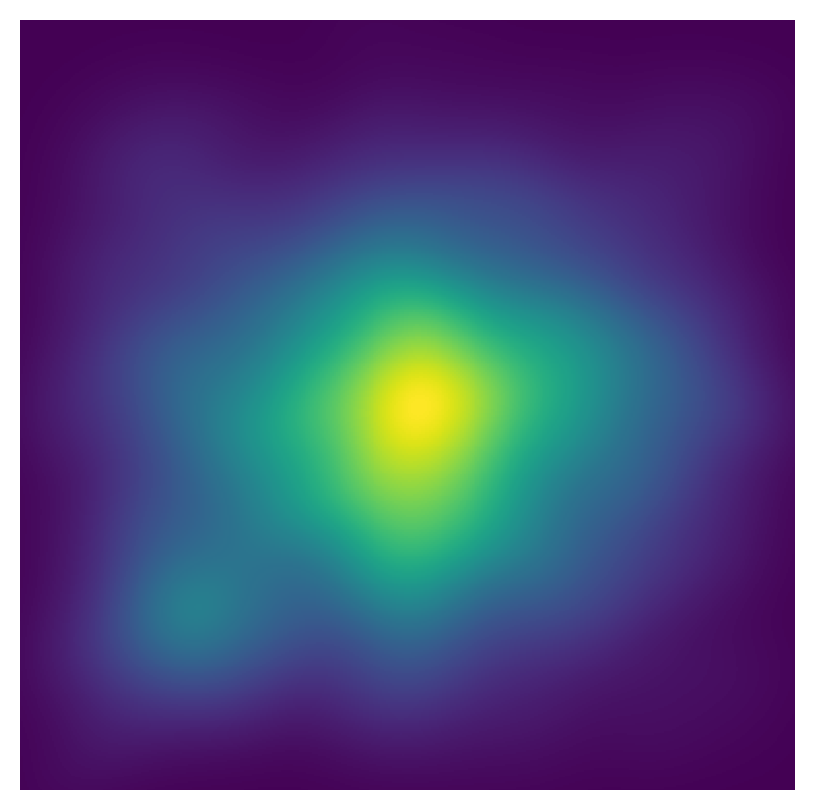}
  \end{subfigure}
\begin{subfigure}[t]{0.195\textwidth}
    \centering
    \caption{\textbf{2nd}}
    \includegraphics[width=\linewidth]{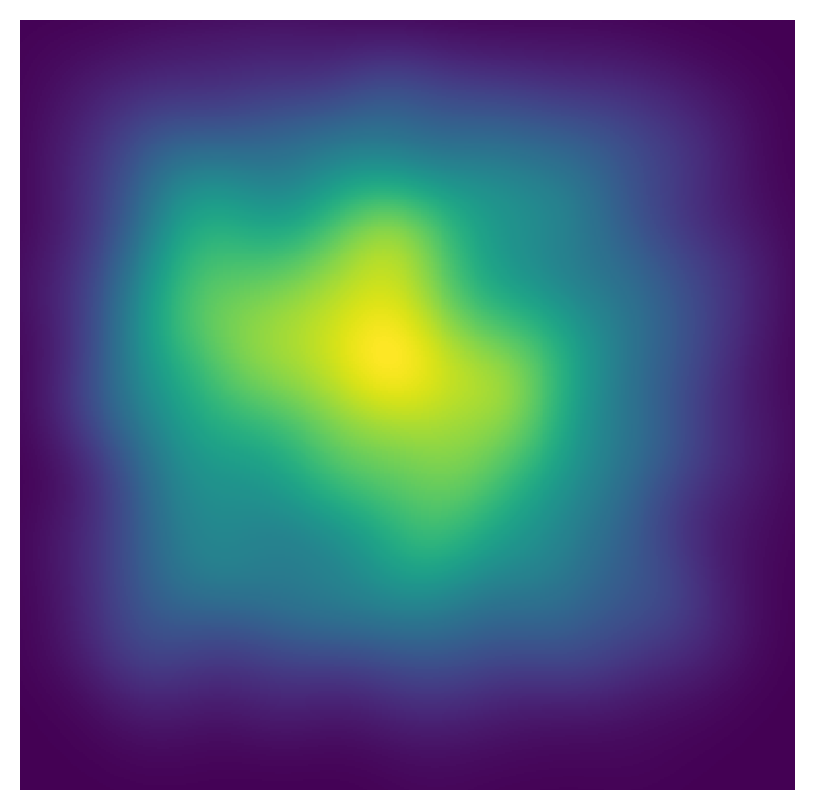}
    \label{fig:webapp_instructions}
  \end{subfigure}
\begin{subfigure}[t]{0.195\textwidth}
    \centering
    \caption{\textbf{3rd}}
    \includegraphics[width=\linewidth]{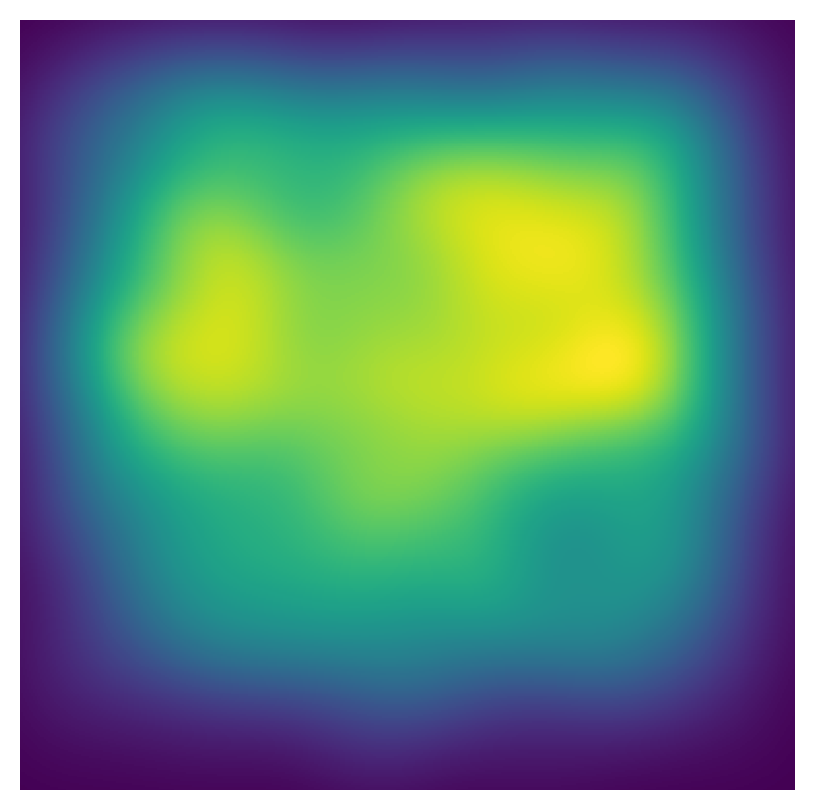}
    \label{fig:webapp_instructions}
  \end{subfigure}
\begin{subfigure}[t]{0.195\textwidth}
    \centering
    \caption{\textbf{4th}}
    \includegraphics[width=\linewidth]{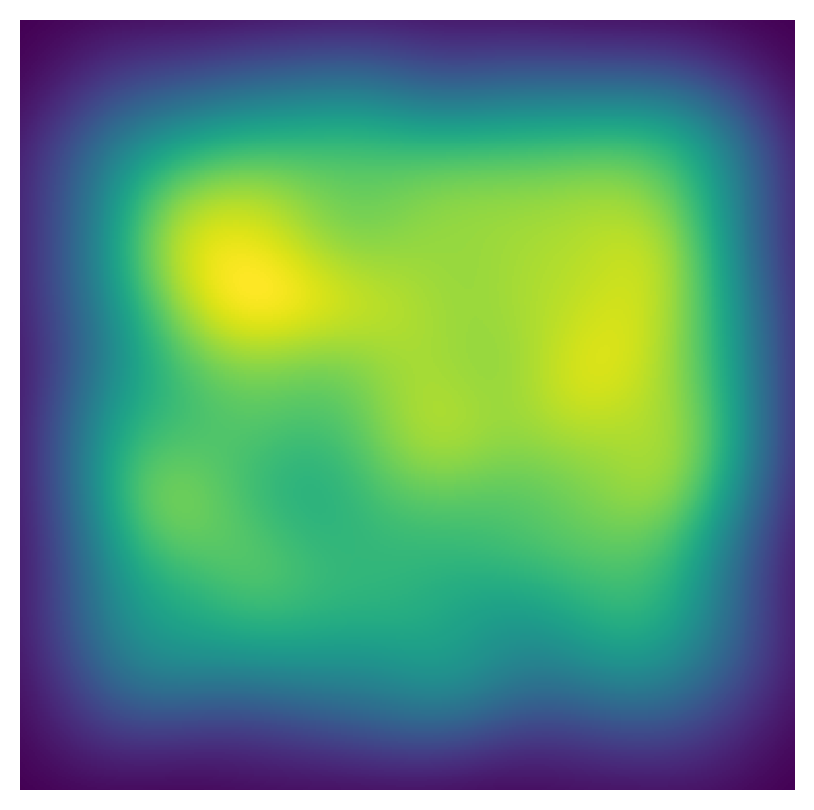}
    \label{fig:webapp_instructions}
  \end{subfigure}
\begin{subfigure}[t]{0.195\textwidth}
    \centering
        \caption{\textbf{5th}}
    \includegraphics[width=\linewidth]{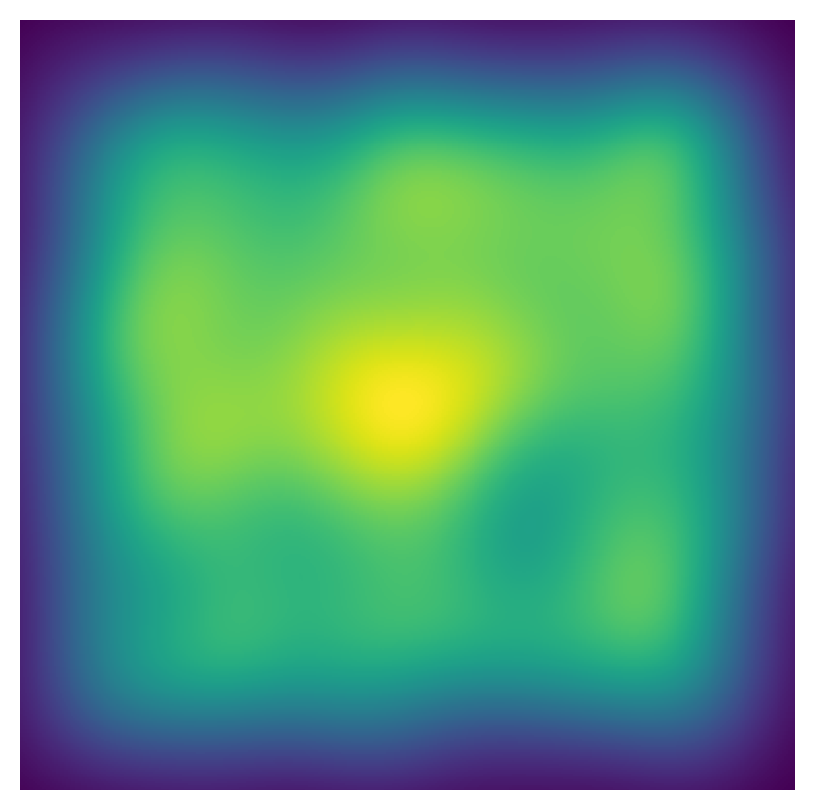}
  \end{subfigure}
  
\begin{subfigure}[t]{0.195\textwidth}
    \centering
    \caption{\textbf{6th}}
    \includegraphics[width=\linewidth]{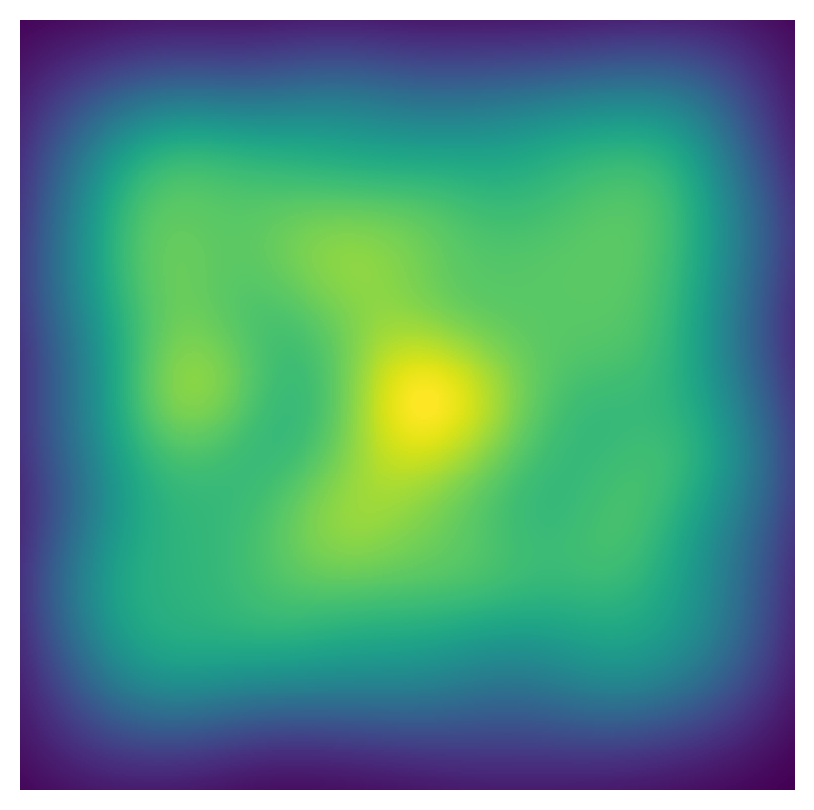}
    \label{fig:webapp_instructions}
  \end{subfigure}
\begin{subfigure}[t]{0.195\textwidth}
    \centering
    \caption{\textbf{7th}}
    \includegraphics[width=\linewidth]{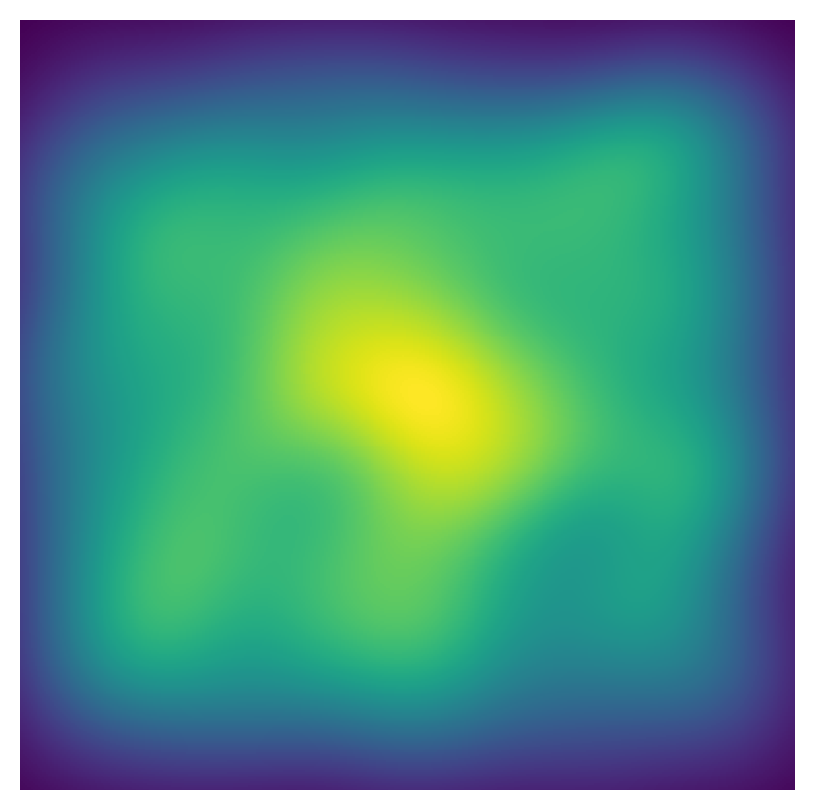}
    \label{fig:webapp_instructions}
  \end{subfigure}
\begin{subfigure}[t]{0.195\textwidth}
    \centering
    \caption{\textbf{8th}}
    \includegraphics[width=\linewidth]{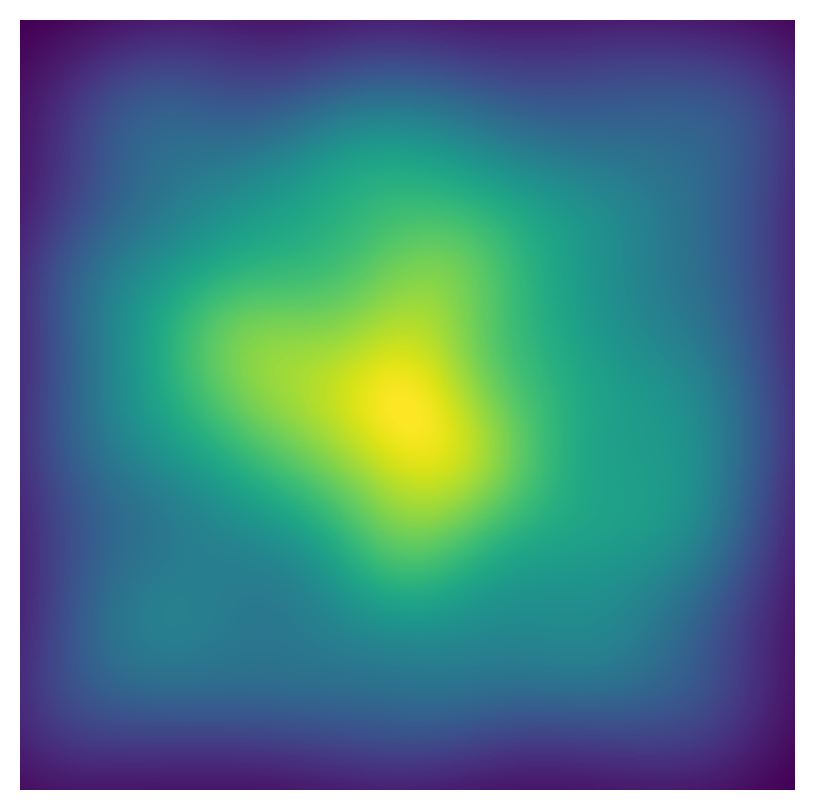}
  \end{subfigure}
\begin{subfigure}[t]{0.195\textwidth}
    \centering
    \caption{\textbf{9th}}
    \includegraphics[width=\linewidth]{imgs/nth_click/8th_click_density.png}
    \label{fig:webapp_instructions}
  \end{subfigure}
\begin{subfigure}[t]{0.195\textwidth}
    \centering
    \caption{\textbf{10th}}
    \includegraphics[width=\linewidth]{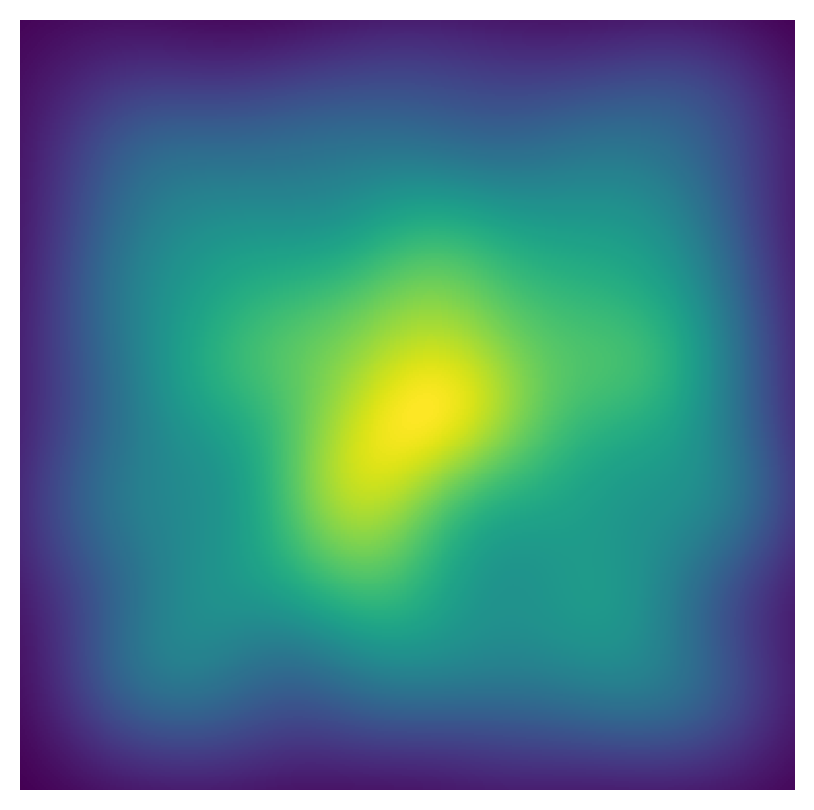}
  \end{subfigure}
  
  \caption{Clicks densities for each of the ten possible clicks.}
  \label{fig:nth_cklick}

\end{figure}

Figure \ref{fig:nth_cklick} shows the estimated clicks densities split by their order. The first two klicks show a strong bias towards the center of the image. Following clicks are often used to perform exploration within the image and the final clicks again show a bias towards the center of the image. 

\newpage
\section{Scanpath examples}
In the subsequent analysis, we present a collection of qualitative examples showcasing scanpaths derived from NevaClip. For each scanpath, both the corresponding captions and a comparison with human reference are provided. This presentation aims to illustrate the effectiveness and reliability of NevaClip in generating scanpaths by examining instances with both accurate and randomly generated captions.

\begin{figure}[htbp]
  \centering
  \begin{subfigure}{0.3\textwidth}
    \includegraphics[width=\linewidth]{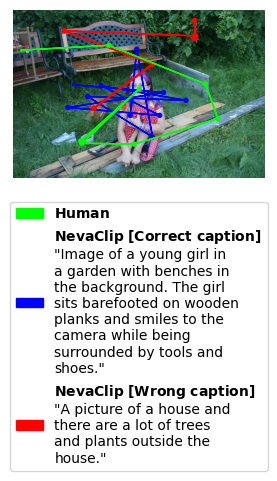}
  \end{subfigure}
  \begin{subfigure}{0.3\textwidth}
    \includegraphics[width=\linewidth]{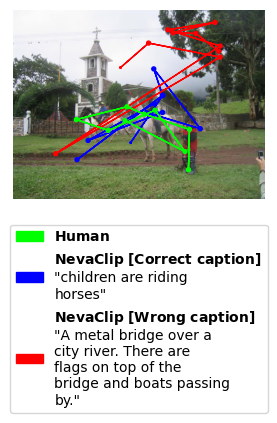}
  \end{subfigure}
  \begin{subfigure}{0.3\textwidth}
    \includegraphics[width=\linewidth]{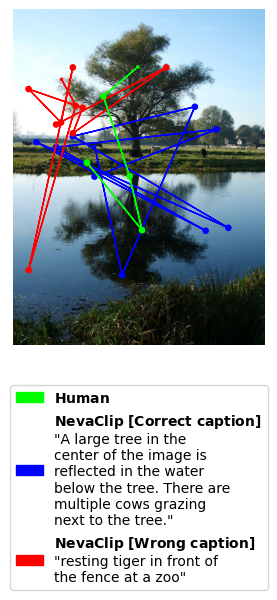}
  \end{subfigure}
  
  \begin{subfigure}{0.3\textwidth}
    \includegraphics[width=\linewidth]{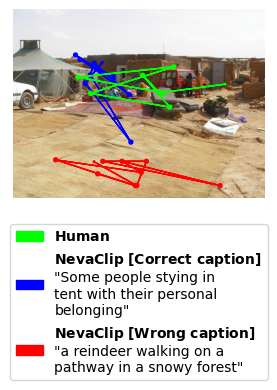}
  \end{subfigure}
  \begin{subfigure}{0.3\textwidth}
    \includegraphics[width=\linewidth]{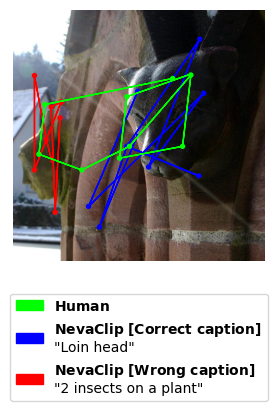}
  \end{subfigure}
  \begin{subfigure}{0.3\textwidth}
    \includegraphics[width=\linewidth]{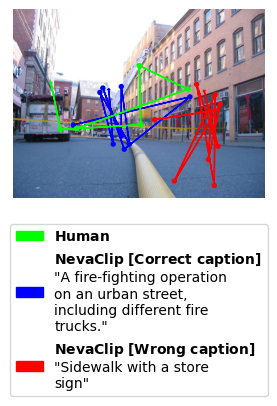}
  \end{subfigure}

\end{figure}
\begin{figure}[htbp]
  \begin{subfigure}{0.3\textwidth}
    \includegraphics[width=\linewidth]{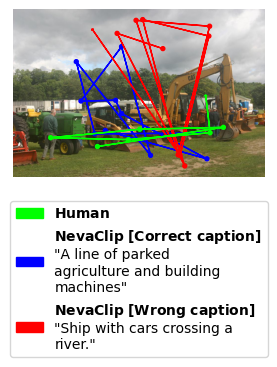}
  \end{subfigure}
  \begin{subfigure}{0.3\textwidth}
    \includegraphics[width=\linewidth]{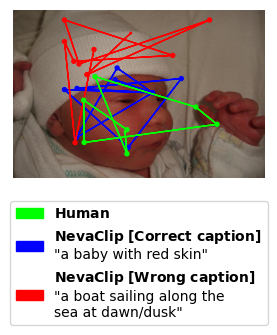}
  \end{subfigure}
  \begin{subfigure}{0.3\textwidth}
    \includegraphics[width=\linewidth]{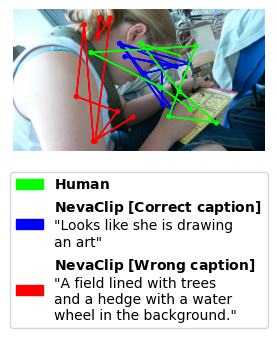}
  \end{subfigure}
  
  \begin{subfigure}{0.3\textwidth}
    \includegraphics[width=\linewidth]{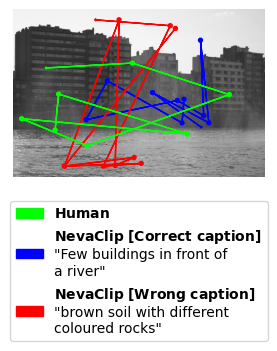}
  \end{subfigure}
  \begin{subfigure}{0.3\textwidth}
    \includegraphics[width=\linewidth]{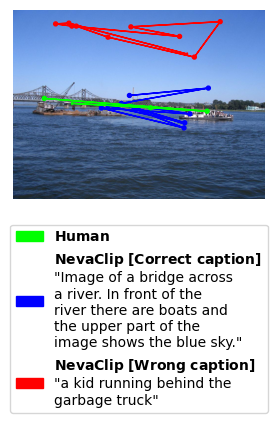}
  \end{subfigure}
  \begin{subfigure}{0.3\textwidth}
    \includegraphics[width=\linewidth]{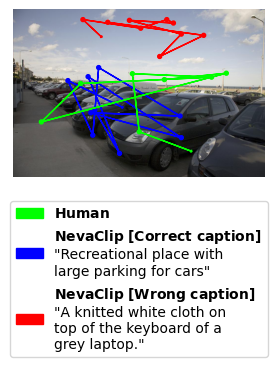}
  \end{subfigure}
  
  \begin{subfigure}{0.3\textwidth}
    \includegraphics[width=\linewidth]{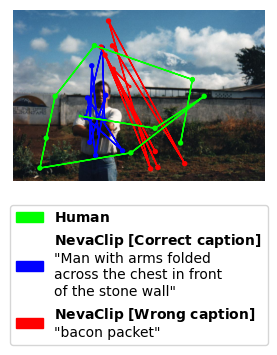}
  \end{subfigure}
  \begin{subfigure}{0.3\textwidth}
    \includegraphics[width=\linewidth]{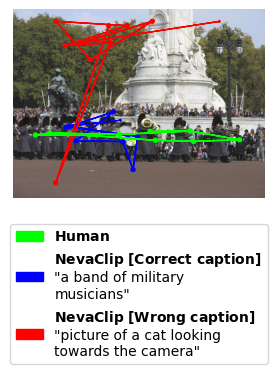}
  \end{subfigure}
  \begin{subfigure}{0.3\textwidth}
    \includegraphics[width=\linewidth]{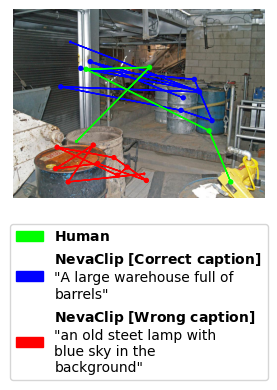}
  \end{subfigure}
  
\end{figure}
\begin{figure}[htbp]
  \begin{subfigure}{0.3\textwidth}
    \includegraphics[width=\linewidth]{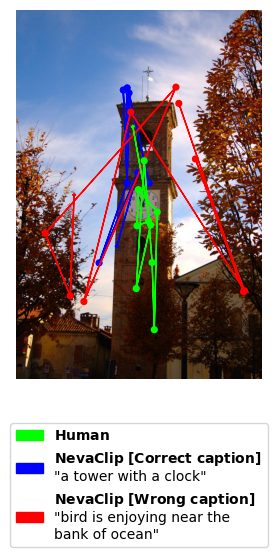}
  \end{subfigure}
  \begin{subfigure}{0.3\textwidth}
    \includegraphics[width=\linewidth]{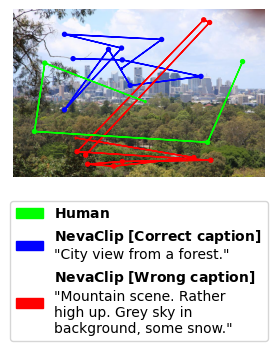}
  \end{subfigure}
  \begin{subfigure}{0.3\textwidth}
    \includegraphics[width=\linewidth]{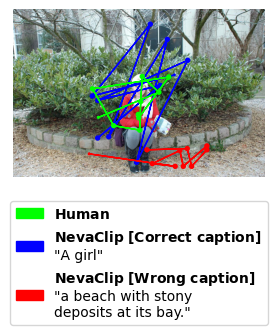}
  \end{subfigure}
  
  \begin{subfigure}{0.3\textwidth}
    \includegraphics[width=\linewidth]{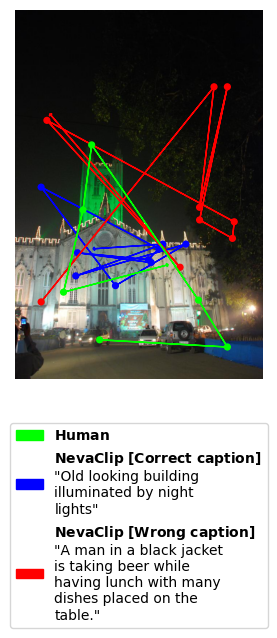}
  \end{subfigure}
  \begin{subfigure}{0.3\textwidth}
    \includegraphics[width=\linewidth]{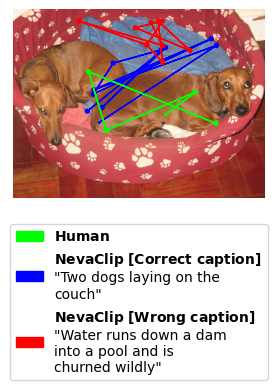}
  \end{subfigure}
  \begin{subfigure}{0.3\textwidth}
    \includegraphics[width=\linewidth]{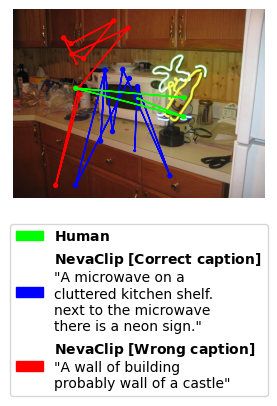}
  \end{subfigure}
  
  \begin{subfigure}{0.3\textwidth}
    \includegraphics[width=\linewidth]{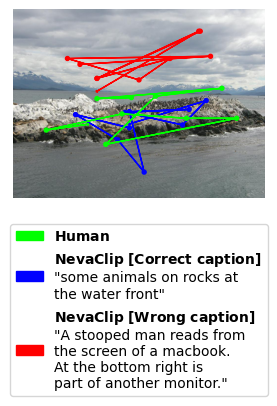}
  \end{subfigure}
  \begin{subfigure}{0.3\textwidth}
    \includegraphics[width=\linewidth]{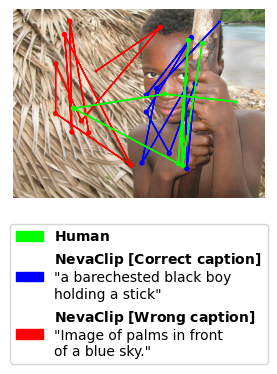}
  \end{subfigure}
  \begin{subfigure}{0.3\textwidth}
    \includegraphics[width=\linewidth]{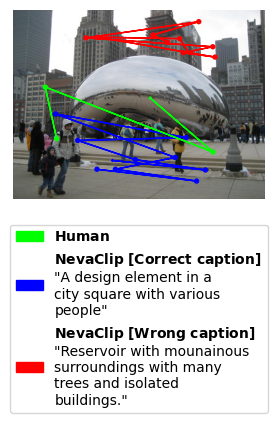}
  \end{subfigure}
  
\end{figure}
\begin{figure}[htbp]
  \begin{subfigure}{0.3\textwidth}
    \includegraphics[width=\linewidth]{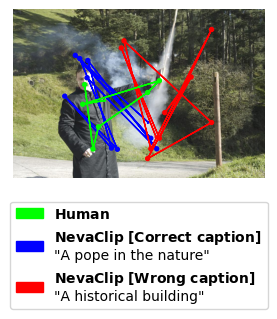}
  \end{subfigure}
  \begin{subfigure}{0.3\textwidth}
    \includegraphics[width=\linewidth]{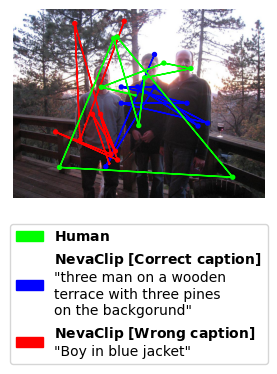}
  \end{subfigure}
  \begin{subfigure}{0.3\textwidth}
    \includegraphics[width=\linewidth]{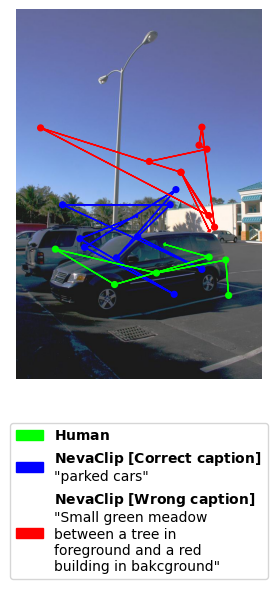}
  \end{subfigure}

  \begin{subfigure}{0.3\textwidth}
    \includegraphics[width=\linewidth]{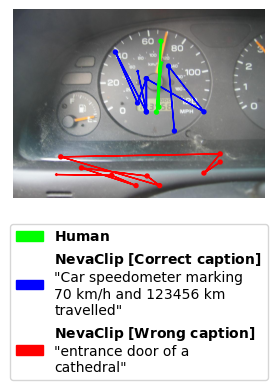}
  \end{subfigure}
  \begin{subfigure}{0.3\textwidth}
    \includegraphics[width=\linewidth]{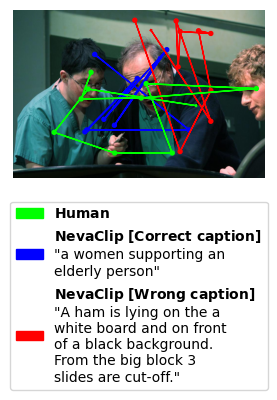}
  \end{subfigure}
  \begin{subfigure}{0.3\textwidth}
    \includegraphics[width=\linewidth]{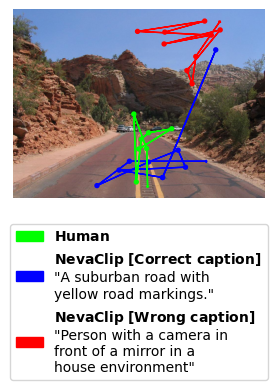}
  \end{subfigure}
  
  \begin{subfigure}{0.3\textwidth}
    \includegraphics[width=\linewidth]{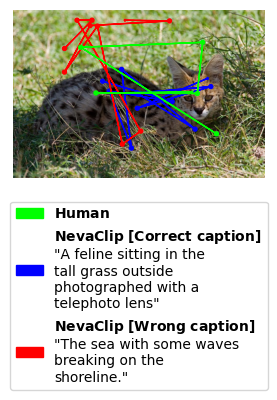}
  \end{subfigure}
  \begin{subfigure}{0.3\textwidth}
    \includegraphics[width=\linewidth]{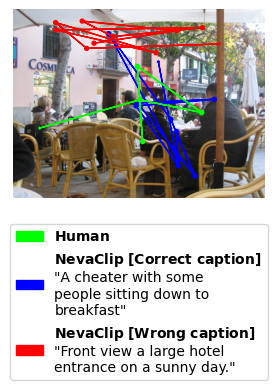}
  \end{subfigure}
  
\end{figure}

\end{document}